\documentclass[10pt,letterpaper]{article}
\usepackage[letterpaper,margin=1in]{geometry}
\usepackage{newtxtext}
\usepackage{natbib}
\setcitestyle{authoryear,round,citesep={;},aysep={,},yysep={;}}
\usepackage{amsmath,amsfonts,bm}

\def\eqref#1{equation~\ref{#1}}

\def\1{\bm{1}}

\DeclareMathAlphabet{\mathsfit}{\encodingdefault}{\sfdefault}{m}{sl}
\SetMathAlphabet{\mathsfit}{bold}{\encodingdefault}{\sfdefault}{bx}{n}

\usepackage[T1]{fontenc}
\usepackage[utf8]{inputenc}
\usepackage{microtype}

\usepackage{amssymb}
\usepackage{amsthm}
\usepackage{newtxmath}

\usepackage{graphicx}
\usepackage{booktabs}
\usepackage{multirow}
\usepackage{hyperref}
\hypersetup{pdftitle={Augment the Cache, Replace the Activation: A 2x2 Study of Internal-State Sharing},pdfauthor={Adam Deng, Shicheng Liu, Yada Zhu}}
\usepackage{url}

\newtheorem{proposition}{Proposition}
\newtheorem{conjecture}{Conjecture}

\newcommand{\lead}[1]{\textbf{#1}}

\newcommand{\Kc}{\mathbf{K}}
\newcommand{\Vc}{\mathbf{V}}
\newcommand{\hs}{\mathbf{h}}
\newcommand{\Cnat}{C^{\star}}
\newcommand{\Rd}{\mathcal{R}}
\newcommand{\Tol}{\mathcal{T}}

\title{Augment the Cache, Replace the Activation:\\
A 2x2 Study of Internal-State Sharing}

\author{Adam Deng\\MIT ORC \and Shicheng Liu\\MIT CSAIL \and Yada Zhu\\IBM Research}

\date{}

\begin{document}

\maketitle

\begin{abstract}
Multi-agent systems built from frozen language models communicate through text, which discards the internal state the sender already computed and makes the receiver pay to rebuild it. Work on internal channels avoids that cost, framing the problem as \emph{transfer}: agent A sends information to agent B. Every agent pair that must communicate needs its own transfer bridge, and every agent added to a system needs a new bridge fitted against each previous agent. We reframe this problem as sharing: an agent publishes its internal state once into a pool, and every peer reads that pool through a single mechanism fitted to no sender in particular, so adding an agent adds a publisher and no new bridge. Whereas the carrier of the internal state and the fusion mechanism are usually chosen independently in the literature, our main finding is that these two choices are coupled; the deciding factor is where the carrier enters the receiver’s computation. Consumability is a property of the receiving computation rather than of the representation. We find that the key–value (KV) cache must be augmented rather than swapped in: the receiver cross-attends over a peer’s cache while keeping its own, and with only a small query-side adapter trained, matches the accuracy it gets from reading the source document. Activations invert this. They are best written straight into the receiver, with no bridge, no adapter and no training, which beats exposing them alongside the receiver’s own, though that advantage rests on knowing where in the receiver the state belongs. The rule is \emph{augment the KV cache, replace the activation}, established under one protocol with a shared control ladder. One adapter, trained once, also serves publishers it never trained on, and a pool of several at once, but only those whose internal state lies close to the receiver’s own.\end{abstract}

\section{Introduction and Related Work}
\label{sec:intro}

A multi-agentic language-model system orchestrates several frozen models to plan, criticize, retrieve and divide labor, and has become a dominant pattern for assembling capable behavior from individually insufficient components~\citep{moa,debate,autogen,camel,metagpt,agentsurvey}. Agents in every deployed system of this kind exchange information through one channel: \emph{natural-language text}. Text is the interface because it is the only representation every model agrees on. It is also the model's \emph{output}, a lossy projection of a much richer internal state, and whatever did not survive that projection into words is unavailable downstream. 

\emph{Internal information sharing} is any mechanism conveying what a model computed without requiring that model's outputs, and two objects inside a transformer are the natural carriers. The \lead{KV cache} is the attention memory a transformer~\citep{attention} builds over a context and consults while generating; the \lead{residual-stream activations} are the hidden states $\hs_\ell$ flowing between blocks. Both are computed during reading, both hold the sender's digestion of the context, and neither is an output.

\lead{Sharing, not transfer.} The pairwise mechanisms of prior work on internal channels~\citep{c2c,latentkv,statebridge} can be tuned to a known sender; the shared pool we study cannot, which is the stronger requirement. The appendix tests it against a pool of eight holders whose answer-bearer is unknown (Appendix~\ref{sec:multi}) and against senders never trained on (Appendix~\ref{sec:heldout}).

\lead{Carrier-level channels.} The KV cache is the object the field reached for first, whether as a cross-model exchange channel~\citep{c2c,latentkv,droidspeak,kvcomm,tokendance}, as a \emph{private} object to be shrunk~\citep{h2o,scissorhands,streamingllm,snapkv,minicache,yoco,kvquant,kivi}, or as something to reuse across requests of the \emph{same} model~\citep{pagedattention,promptcache,ragcache}. The latter two families perturb a cache the receiver itself would have produced, which makes cross-\emph{model} reuse a different problem, and one that repeatedly fails without training. Writing into the residual stream is the parallel literature, where much of a model's behavior turns out to live in a few activation directions one can add~\citep{funcvec,icv,repe,iti,caa,actadd,latentsteer,taskarith,modelsoups,ties,fusellm}. That line establishes that \emph{behavior} is portable. We ask whether \emph{content} is. A mature literature on representational correspondence~\citep{cka,svcca,relreps,stitching,lenc2015,gitrebasin,rosetta,platonic,vec2vec} carries one caveat throughout: high representational \emph{similarity} does not imply \emph{interchangeability}. Work that adds information through attention sidesteps the question~\citep{retro,fid,flamingo,perceiver,lora}. In our taxonomy these are all augmentation, and their empirical success is a clue we take seriously.

\lead{Latent communication between agents.} Closest to our framing is a line asking whether agents can exchange latent state in place of text. A survey of eighteen such methods organizes them on an axis of carriers (embeddings, hidden states, KV caches) matching ours, and concludes that a method's carrier and its fusion mechanism ``can be chosen largely independently''~\citep{latentsurvey}. That conclusion is the one this paper contradicts, and the survey's own tabulation pairs the two opposite to our rule on both carriers. Trainable latent channels report gains on reasoning benchmarks~\citep{diffmas}, but two results temper the optimism: a controlled study finds the latent channel matches yet never exceeds text~\citep{latentlimits}, and a causal audit shows aggregate accuracy does not identify the latent message's contribution, so agents can appear to communicate while the channel carries little~\citep{latentaudit}. Our control ladder (Section~\ref{sec:methods}) exists because of the second. StateBridge is the closest positive cross-model result, aligning last-layer hidden states across \emph{different} models with a training-free orthogonal map~\citep{statebridge}. Read our cross-family collapses against it: StateBridge hands \emph{last-layer} states to a downstream consumer, while we inject at \emph{intermediate} layers the receiver's remaining stack must consume.

\subsection{This paper}

\begin{figure}[t]
  \centering
  \includegraphics[width=0.98\linewidth]{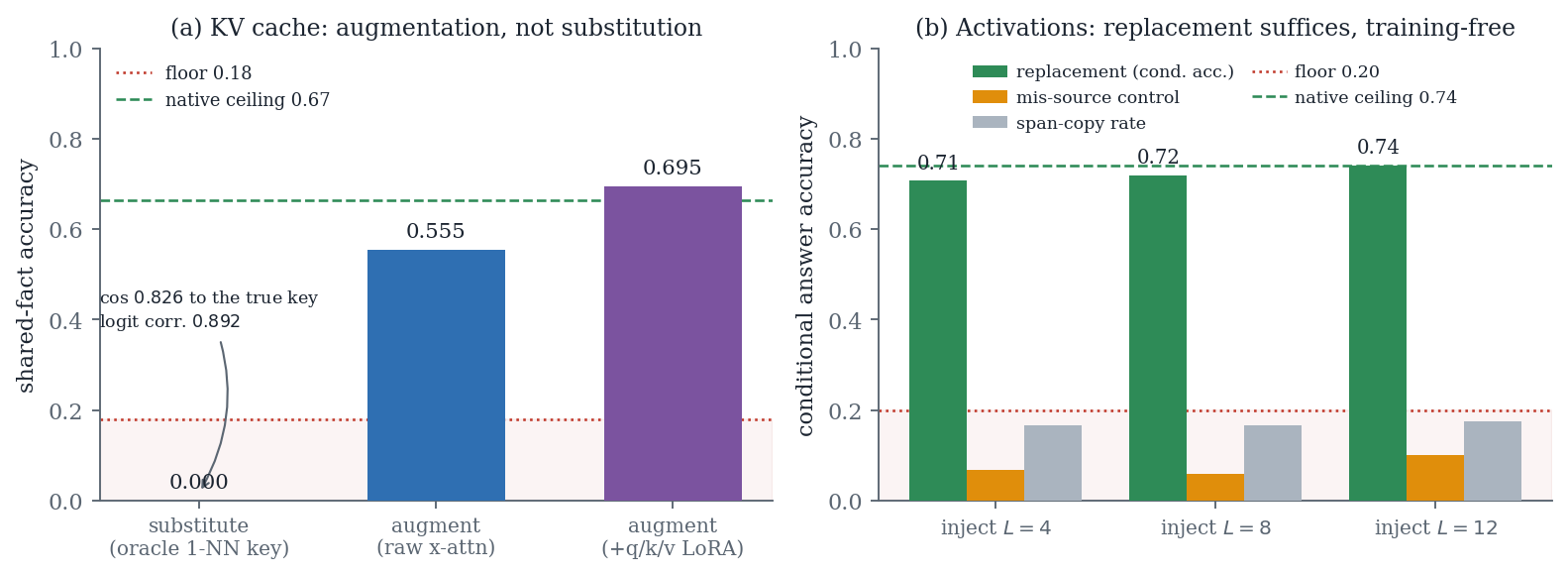}
  \caption{The two carriers take opposite mechanisms. \textbf{(a)} The KV cache cannot be substituted, even at the fidelity annotated on the bar, and must be augmented instead. \textbf{(b)} Activations invert this: direct replacement works, training-free. Floors and native references are per carrier.}
  \label{fig:tworegimes}
\end{figure}

We study both carriers under one controlled protocol, and Figure~\ref{fig:tworegimes} states the result in one picture. Our contributions are as follows.

\lead{(1) A receiver-centric frame.} We define a receiver's \emph{readout map} and the \emph{tolerable set} it induces (Section~\ref{sec:theory}), separating two regimes the literature routinely conflates: under \emph{substitution} a shared object replaces the receiver's own, under \emph{augmentation} the receiver keeps its own and the shared object is exposed alongside it. That distinction is what makes the KV and activation results commensurable.

\lead{(2) A mechanism for why KV substitution fails at high fidelity.} The asymmetry follows from where each object is consumed (Section~\ref{sec:theory}): keys and values are read \emph{by} the softmax of Eq.~\ref{eq:attn}, a normalized competitive map, while activations are read before any softmax through a near-unit-gain path. Our measured readout gain falsifies the conditioning cliff such an argument invites, so the bound is not a matter of numerical sensitivity (Section~\ref{sec:kv}).

\lead{(3) The scope of the substitution bound.} It reproduces on seven receivers from three families (Appendix~\ref{sec:arch}), and it holds in the most favorable case physically available, where the substituted key is the nearest one the receiver ever produced itself (Section~\ref{sec:kv}). Every cross-family \emph{sharing} configuration we test, on either carrier, collapses to the floor.

\lead{(4) A 2x2 grid evaluation of carrier (KV cache, activation)-mechanism (replace, augment).} KV augmentation clears the floor before anything is trained, so a usable channel exists without fitting an alignment map at all (Section~\ref{sec:kv}). Activation replacement needs no training whatsoever, but it requires oracle position and layer alignment that our same-base setting supplies and a deployed system would have to establish (Section~\ref{sec:acts}).

\section{A receiver-centric account of consumability}
\label{sec:theory}

This section fixes the objects and defines the two regimes. It then gives the conditioning argument that predicts the asymmetry we go on to measure.

\subsection{Receivers, carriers, and readout maps}

Fix a frozen transformer $M$ with $L$ layers, hidden width $d$, and grouped-query attention~\citep{gqa} with $N_Q$ query heads over $N_{KV}$ key/value heads in groups of $G = N_Q/N_{KV}$, head dimension $d_h$. Frozen means its weights, and so every function it applies, stay fixed throughout. Nothing below fine-tunes the receiver.

Running $M$ on a context $x$ produces, at every layer $\ell$ and position $t$, a residual-stream activation $\hs_\ell(t) \in \mathbb{R}^{d}$ and the derived attention memory \[ \Kc_\ell(t) = \mathrm{RoPE}\!\left(W_K^{(\ell)}\,\mathrm{LN}(\hs_\ell(t))\right), \qquad \Vc_\ell(t) = W_V^{(\ell)}\,\mathrm{LN}(\hs_\ell(t)). \] A \emph{carrier} is the object a sharing mechanism transports. We consider three: the activation $\hs$, and the cache components $\Kc$ and $\Vc$. The dependency matters. $\Kc$ and $\Vc$ are functions of $\hs$ through the \emph{producer's own} projections, which welds the cache to the geometry of the model that produced it in a way the activation is not.

Let $\mathcal{C}$ be the space of caches. The receiver's \emph{readout map} $\Rd : \mathcal{C} \to \Delta(V)$ sends a cache $C$ to the next-token distribution $M$ produces when its attention consumes $C$: the composition of the attention-over-$C$ operation, the residual stream, the remaining blocks, the final norm and the unembedding, all with $M$'s frozen weights. $\Rd$ is \emph{receiver-specific}: the same $C$ fed to two receivers passes through two different maps. There is no reason $\Rd(C)$ should mean anything to a receiver other than the one whose forward pass $\Rd$ encodes.

One cache stands apart from the rest: the \emph{native cache} $\Cnat$, which $M$ itself produces on the true context. $\Rd(\Cnat)$ is the receiver reading its own work, and it defines the \emph{native ceiling}.

For a behavioral divergence $D$ on next-token distributions and a tolerance $\epsilon$,
\begin{equation}
\Tol(M,\epsilon) \;=\; \bigl\{\, C \in \mathcal{C} \;:\; D\bigl(\Rd(C),\, \Rd(\Cnat)\bigr) \le \epsilon \,\bigr\}.
 \label{eq:tolerable}
\end{equation}
$\Tol$ is the set of caches the receiver consumes behaviorally-equivalently to its own.

\subsection{Two regimes: substitution and augmentation}

The literature on cache sharing mixes two mechanisms that this frame separates cleanly. That separation carries most of the explanatory weight in what follows.

A \emph{substitution} mechanism is a map $g$ producing an object $g(\text{source})$ that is placed \emph{in the receiver's own slots}, so the receiver attends over $g(\text{source})$ \emph{instead of} its own state. Reconstruction, selection from a bank, quantization, and direct overwrite are all substitution. Its behavior is $\Rd(g(\text{source}))$, so Eq.~\ref{eq:tolerable} governs it: substitution succeeds only if $g$'s image intersects $\Tol$.

An \emph{augmentation} mechanism leaves the receiver's own state intact and exposes the source additionally: the receiver attends over the concatenation $[\Cnat; C_{\mathrm{src}}]$, or receives the source as an additive residual contribution. Here the receiver never applies $\Rd$ to a foreign point, so Eq.~\ref{eq:tolerable} does not bound it. The receiver now has strictly more to attend over than its own context alone, and augmentation can in principle exceed $\Rd(\Cnat)$.

\subsection{Why the carriers differ: the consumption point}

What decides the outcome is \emph{where} each carrier enters the receiver's computation.

An activation is consumed by the next block through $\hs \mapsto \hs + f(\mathrm{LN}(\hs))$: a normalization, a linear map, an \emph{additive residual}. Gain along that path is $O(1)$, so a displacement $\delta$ passes through essentially undistorted while the normalization re-projects toward the operating manifold at every layer.

A cache is consumed by exactly one operation,
\begin{equation}
  \mathrm{Attn}(q, \Kc, \Vc) \;=\;
  \mathrm{softmax}\!\left(\frac{q^{\top}\Kc}{\sqrt{d_h}}\right)\Vc ,
  \label{eq:attn}
\end{equation}
and the softmax is \emph{normalized and competitive}: it scores each key against every other key, concentrating weight on whichever keys lead. A displacement leaving a key almost exactly where it was in Euclidean or cosine terms can still change the \emph{ordering} of the $q^{\top}k$ scores, and once the ordering changes, weight moves to different tokens. Consumability turns on whether $\delta$ disturbs the narrow margins by which the true keys presently win. Some perturbations disturb none, so $\Tol$ is not literally zero-dimensional: a common shift in all logits, or a displacement confined to keys the query never ranks highly, leaves the readout intact. Our claim is weaker and empirical. For the substitutions we can construct, the usable neighborhood around $\Cnat$ is narrow enough that no mechanism we test finds a point inside it, which we read as measured tolerance rather than the geometry of $\Tol$.

Two consequences follow. Both run against intuition, and we test both.

\begin{proposition}[Fidelity does not imply consumability]
\label{prop:decoupling}
Let $\phi(C)$ be any geometric fidelity of $C$ to $\Cnat$ (cosine of the key/value tensors, MSE, or correlation of induced attention logits). High $\phi$ does not imply $C \in \Tol$. Cosine measures Euclidean alignment while the softmax depends only on the \emph{ranking} of $q^{\top}k$, and a cosine-invisible rotation can reorder that ranking. A substitution map may drive $\phi$ arbitrarily close to its maximum while $D(\Rd(C),\Rd(\Cnat))$ stays pinned at its worst.
\end{proposition}

\begin{proposition}[The wall is not amplification, and not depth]
\label{prop:notdepth}
Write $\kappa_{\mathrm{off}}$ and $\kappa_{\mathrm{on}}$ for the readout Jacobian gain under off- and on-manifold perturbation. A conditioning cliff would require $\kappa_{\mathrm{off}} \gg \kappa_{\mathrm{on}}$; depth-compounding would kill a carrier traversing the same $L$ layers alongside the cache. Neither holds (Appendix~\ref{sec:results-mech}): the measured ratio is close to unity, and activations injected into the identical $L{=}28$ receiver transfer while caches do not. The wall belongs to the readout a carrier feeds rather than the depth it is read through.
\end{proposition}

One analytic bound sharpens the GQA dependence. Because $G$ query heads share each key/value slot, a perturbation there is read out through all $G$ queries and the per-slot sensitivity constant scales as $\sqrt{2G}$; for $G{=}6$ this is $\sqrt{12} \approx 3.46$. The constant belongs to attention \emph{shape} rather than the weights, and at $O(1)$ it is far too small to build a wall by amplification (Proposition~\ref{prop:notdepth}).

\subsection{The Asymmetry Conjecture}

The frame yields a falsifiable prediction, and that prediction organizes the rest of the paper.

\begin{conjecture}[Asymmetry of internal information sharing]
\label{conj:asym}
For a frozen receiver,
\begin{enumerate}
  \item[(C1)] \textbf{The cache must be augmented.} Because the usable neighborhood around $\Cnat$ is narrow, any substitution of keys and values is bounded near the transfer floor, whatever fidelity $\phi$ it achieves. Augmentation never asks the receiver to consume a foreign cache in place of $\Cnat$, so $\Tol$ does not bound it, and it can approach or exceed the native ceiling.
  \item[(C2)] \textbf{The activation may be replaced.} Because the activation is
consumed through a near-unit-gain pre-softmax path, its tolerable region is a genuine neighborhood rather than a point, and direct replacement transfers. Augmenting the same carrier only tilts the receiver's own state instead of installing the holder's, and it transfers \emph{less}.
\end{enumerate}
The conjecture is falsified if a cache substitution transfers well above the floor while provably outside $\Tol$; if cache augmentation with query-side adaptation fails to clear the floor; or if activation replacement fails while activation augmentation succeeds.
\end{conjecture}

\section{Methodology}
\label{sec:methods}

\lead{Models.} Unless noted, holder and receiver are the same frozen Qwen2.5-1.5B~\citep{qwen2.5} with $L{=}28$ layers, hidden width $1536$, and grouped-query attention in groups of $G{=}6$. The cross-family holder is GPT-Neo-1.3B, mapped into the receiver's key space by a frozen per-layer SVD projection. Weights are frozen in every experiment; only bridge parameters, where present, are trained. Appendix~\ref{app:methods-detail} gives the full geometry of both models.

\lead{The split-context fact-transfer probe.} Every number in this paper comes from one instrument. An item is a SQuAD~\citep{squad} passage whose answer appears verbatim and runs to at most four words, paired with its question, in two versions: \emph{full}, the original passage, and \emph{masked}, the passage with the answer span replaced by \texttt{\_\_\_\_}. The \emph{holder} processes the full passage. The frozen \emph{receiver} gets the masked passage, so it cannot answer from its own context. We then supply the receiver an object derived from the holder ($\Kc$, $\Vc$, or $\hs$) and ask whether it can now answer. Prompts are \texttt{\{passage\}$\backslash$nQuestion:} \texttt{\{q\}$\backslash$nAnswer:}, truncated to $256$ tokens, with $8$ greedy answer tokens. Evaluation items come from the validation split, training items from train.

\lead{Anchors.} Two per-receiver anchors bound every measurement. The \emph{floor} is the receiver on the masked passage with no useful injection: it knows the question and has a fact-free context. The \emph{native reference} is the receiver on the full passage reading its own state, the identity case $\Rd(\Cnat)$. \emph{The KV-side and activation-side probes carry different anchors and must not be compared in absolute terms.} KV-side runs score floor $0.18$ / ceiling $0.665$ at $n{=}200$, activation-side runs $0.20$ / $0.742$ at $n{=}120$. We report position within the floor--ceiling band and plot both anchors on every panel.

\lead{Metrics.} Generation is scored by normalized substring match: the gold answer, lowercased and stripped of punctuation, must appear in the generation. We report \lead{shared-fact accuracy} over all items. Because a receiver might answer by copying a visible span, we also report \lead{conditional answer accuracy}, restricted to items the receiver answers correctly at its ceiling, beside an explicit \lead{span-copy rate} (the answer shares a two-token contiguous bigram with the injected span \emph{and} was produced). A transfer claim is credited only when shared accuracy clears the floor \emph{and} conditional accuracy exceeds span-copy by a wide margin.

\lead{Controls.} Four, of increasing strength, defined in Appendix~\ref{app:methods-detail}: a \emph{no-op} injecting the receiver's own masked state at the same sites, a \emph{position} control at the fact-free prefix, a \emph{mis-source} control supplying a different item's state, and an \emph{answer-disjoint split} for trained bridges. The last two separate genuine routing from a mechanism emitting memorised answers.

\lead{Mechanisms compared.} On the KV side, \emph{substitution} places produced or selected keys in the receiver's own slots, while \emph{augmentation} keeps the receiver's cache and cross-attends over the concatenation with the holder's, used verbatim. Only the receiver's own projections adapt, by low-rank adapters that start at exactly the identity~\citep{lora}. On the activation side, \emph{replacement} overwrites $\hs_\ell$ at the fact region, training-free. One caveat belongs here and not later. The fact region comes from diffing the masked token sequence against the full one, and the injection layer is chosen from a swept set, so position and depth are both supplied to the mechanism. That is an oracle, and it makes the activation arm a measurement of consumability under correct alignment and not a deployable interface (Appendix~\ref{app:methods-detail}). Activation \emph{augmentation} instead adds a routed convex combination of source activations to the residual stream.

\lead{Verification.} Attention is re-implemented manually to permit injection, so every run must first reproduce native greedy decoding string-exactly and rebuild the native cache to within tolerance before its numbers are read (Appendix~\ref{app:methods-detail}).

\section{Experimental Results and Analysis}
\label{sec:results}

\subsection{The KV cache: substitution is floor-bounded}
\label{sec:kv}

We first establish the bound of Conjecture~\ref{conj:asym}(C1) under conditions engineered to be maximally favorable to substitution. \emph{Synthesizing} a key risks landing off-manifold, so instead we \emph{select} one from a bank of $6000$ genuine keys the receiver itself produced on other contexts. Every candidate is then on the receiver's own key manifold by construction. For a target slot we take the $k$-th nearest neighbor of the true key and substitute it.

The result is categorical. Full substitution by the nearest genuine key transfers at $0.000$. The tolerance is flat in neighbor rank ($k{=}1,\dots,16$ all $\le 0.01$), so there is no graded neighborhood to exploit. Interpolating from the true key toward the neighbor craters: every cell at mixing weight $\alpha \ge 0.5$ sits at or below $0.02$ (Table~\ref{tab:oracle}). The \emph{identity} arm, which hands the receiver its own exact keys, reproduces the ceiling $0.665$ bit-exactly, with cosine $1.000$ and all $28$ layers clearing the fidelity threshold. So the correct point exists, is on-manifold, and is fully consumable. The \emph{nearest available miss} already returns nothing.

\begin{table}[t]
\label{tab:oracle}
\begin{center}
\begin{tabular}{lcccccc}
\toprule
 & $\alpha{=}0.0$ & $0.1$ & $0.25$ & $0.5$ & $0.75$ & $1.0$ \\
\midrule
$k{=}1$  & 0.180 & 0.185 & 0.110 & 0.005 & 0.000 & 0.000 \\
$k{=}2$  & 0.180 & 0.180 & 0.110 & 0.020 & 0.010 & 0.005 \\
$k{=}4$  & 0.180 & 0.205 & 0.120 & 0.010 & 0.005 & 0.000 \\
$k{=}8$  & 0.180 & 0.175 & 0.120 & 0.005 & 0.000 & 0.000 \\
$k{=}16$ & 0.180 & 0.145 & 0.135 & 0.020 & 0.015 & 0.010 \\
\bottomrule
\end{tabular}
\caption{Oracle substitution: the receiver's key is interpolated toward the $k$-th nearest key the receiver itself produced elsewhere, so every candidate is on its own manifold. No neighbor rank recovers a graded tolerance. Qwen2.5-1.5B, $n{=}200$, floor $0.18$, ceiling $0.665$.}
\end{center}
\end{table}

This is Proposition~\ref{prop:decoupling} in its sharpest form. The bank-selected key attains cosine $0.826$ and attention-logit correlation $0.892$ to the true key, which is high fidelity by any conventional proxy, and it transfers nothing. A broader sweep of substitution mechanisms (freeform regression, manifold-matched training objectives, column- and row-space construction, worst-case hinge losses, high-capacity per-layer MLPs, and codebook selection) spans transfer $0.00$ to $0.20$ against the same floor of $0.18$. None clears it. The best-fidelity member of the family is also not the best-transferring one.

\subsection{The KV cache: augmentation clears the ceiling}

We change only the mechanism. The receiver keeps its own real cache and cross-attends over the concatenation with the holder's verbatim cache; only query-side parameters adapt. Table~\ref{tab:xkv} reports the ladder.

\begin{table}[t]
\label{tab:xkv}
\begin{center}
\begin{tabular}{llcccc}
\toprule
Setting & Adapted & Shared & Control & $\Delta$ vs.\ raw & vs.\ ceiling \\
\midrule
\multirow{4}{*}{same model}
 & none (raw)   & 0.555 & 0.185 & ---    & $-0.110$ \\
 & $q$          & 0.600 & 0.155 & $+0.045$ & $-0.065$ \\
 & $k,v$        & 0.640 & 0.090 & $+0.085$ & $-0.025$ \\
 & $q,k,v$      & \textbf{0.695} & 0.200 & $+0.140$ & $+0.030$ \\
\midrule
\multirow{4}{*}{cross family}
 & none (raw)   & 0.010 & 0.005 & ---    & --- \\
 & $q$          & 0.010 & 0.005 & $+0.000$ & --- \\
 & $k,v$        & 0.080 & 0.085 & $+0.070$ & --- \\
 & $q,k,v$      & 0.040 & 0.035 & $+0.030$ & --- \\
\bottomrule
\end{tabular}
\caption{KV augmentation. The receiver keeps its own cache and cross-attends over the holder's verbatim cache; only the receiver's projections adapt. The control feeds a holder cache built on the answer-masked passage. Same-model, $n{=}200$.}
\end{center}
\end{table}

We have three readings. First, \emph{augmentation escapes the substitution bound immediately}: the untrained bridge, with zero trainable parameters, already lands several times clear of the floor, which no substitution mechanism we tested could approach. The mechanism does the work here, not the optimization. Second, the adaptation ladder is monotone, $q,k,v \succ k,v \succ q \succ$ raw, so adapting the receiver's own projections is what converts available signal into answers. Third, the best arm meets the receiver's native reference, and we are deliberately careful about what that licenses. At $n{=}200$ the binomial standard error on the difference is $0.047$, so $0.695$ versus $0.665$ is $z \approx 0.6$ and \emph{not} statistically distinguishable from a tie. The defensible claim is that query-side-adapted augmentation \emph{reaches} the native reference; we cannot claim it provably exceeds it, since establishing that needs multiple seeds we have not run.

\lead{Is this lexical retrieval?} The holder's cache contains the answer passage verbatim and the receiver cross-attends over it, so a high score could in principle come from locating and copying answer tokens instead of consuming the holder's reading. We close this by porting the hardened activation-side ladder across, on an answer-disjoint evaluation where no evaluation answer appears in the bridge's training data. Conditional accuracy clears span-copy by $0.519$ ($z \approx 11$), the reverse of what contamination predicts, and the mis-source arm, handed a \emph{different} item's answer-bearing passage, sits below floor. Widening the pool sharpens this: span-copy \emph{falls} while conditional accuracy persists, though more distractors would give a copying mechanism more to copy. Two caveats. Copying is not eliminated, only shown insufficient as an explanation, since a span-copy rate of $0.366$ at $N{=}1$ means some answers do arrive as lifted spans. And the mis-source arm, clean at $N{=}1$, rises above floor in the larger pools, where the trained bridge recovers some answers whichever source carries them; there we stand behind the control-subtracted lift and not the raw accuracy.

\lead{Which senders, and how many.} Two sweeps bound the word \emph{compatible}. Facing a pool of $N$ holders of which one carries the answer, raw attention drops below floor early while the trained query-side bridge keeps a positive control-subtracted lift out to $N{=}8$ (Appendix~\ref{sec:multi}); varying the publisher instead, a bridge trained on the receiver's own state still transfers at cosine $0.9994$ to the receiver's cache and is indistinguishable from its control by $0.9933$ (Appendix~\ref{sec:heldout}, Figure~\ref{fig:heldout}). Compatibility is narrow and measurable.

\subsection{Activations: replacement transfers, and beats augmenting}
\label{sec:acts}

We switch carrier and keep the probe. Two things are given to this mechanism and should be kept in view while reading the numbers: the fact region, obtained by diffing masked against full, and the injection layer, taken from a swept set. What follows measures how consumable a replaced activation is when it lands in the right place, and not whether an agent could find that place. Overwriting the receiver's residual stream at the fact region with the holder's activations clears the floor at every depth probed and peaks at $L{=}12$, where conditional accuracy is numerically indistinguishable from the receiver's native reference (Table~\ref{tab:acts}). This is training-free: no cross-attention, no bridge, no trained parameters anywhere. The no-op control sits at \emph{exactly} the floor at every layer, and the position control likewise, so the lift is not ``something was overwritten here''. Conditional accuracy exceeds span-copy roughly fourfold, which rules out regurgitation.

\begin{table}[t]
\label{tab:acts}
\begin{center}
\begin{tabular}{lccccc}
\toprule
Inject layer & Shared fact & Cond.\ acc. & Span-copy & Mismatch ctrl & No-op ctrl \\
\midrule
$L{=}4$   & 0.533 & 0.708 & 0.167 & 0.067 & 0.200 \\
$L{=}8$   & 0.550 & 0.719 & 0.167 & 0.058 & 0.200 \\
$L{=}12$  & 0.567 & \textbf{0.742} & 0.175 & 0.100 & 0.200 \\
$L{=}16$  & 0.567 & 0.708 & 0.167 & 0.408 & 0.200 \\
$L{=}20$  & 0.575 & 0.708 & 0.183 & 0.425 & 0.200 \\
all of the above & 0.517 & 0.685 & 0.142 & 0.067 & 0.200 \\
\bottomrule
\end{tabular}
\caption{Activation replacement, training-free. The receiver's hidden states at the fact region are overwritten with the holder's. Qwen2.5-1.5B, $n{=}120$ ($\mathrm{cond}\,n{=}89$), floor $0.20$, ceiling $0.742$. The mismatch control is clean at $L \le 12$ but not at $L \ge 16$ (shaded reading in the text).}
\end{center}
\end{table}

\lead{A caveat we must not bury.} The mismatch control is clean at $L{\le}12$ ($0.058$--$0.100$, below floor) but \emph{not} at $L{\ge}16$, where it rises to $0.408$--$0.425$. Injecting the holder's fact at the \emph{wrong} positions lifts the receiver $\approx 2\times$ above floor at exactly the depths where replacement performs best. The no-op and position controls stay pinned everywhere, so the core claim survives. We read the clean result off the $L{\in}\{4,8,12\}$ band and treat the deeper layers as suggestive only.

\lead{Replacement beats augmentation on this carrier.} We ran the same carrier through an augmentation mechanism instead, adding a routed convex combination of three holders' activations to the receiver's residual stream, again training-free. Its best cell reaches $0.438$, clear of the floor yet far short of direct replacement. The dose response is monotone and gentle-is-better ($\alpha{=}0.5 \succ 1 \succ 2 \succ 4$ at every depth). In line with a robust finding across this program, the router is \emph{inert}: an oracle per-token router that sees the correct source earns nothing over the uniform one. Augmenting only \emph{tilts} the receiver's own state, while replacement \emph{installs} the holder's. Mechanism preference inverts between the two carriers, which is Conjecture~\ref{conj:asym}(C2).

\begin{figure}[t]
  \centering
  \includegraphics[width=0.76\linewidth]{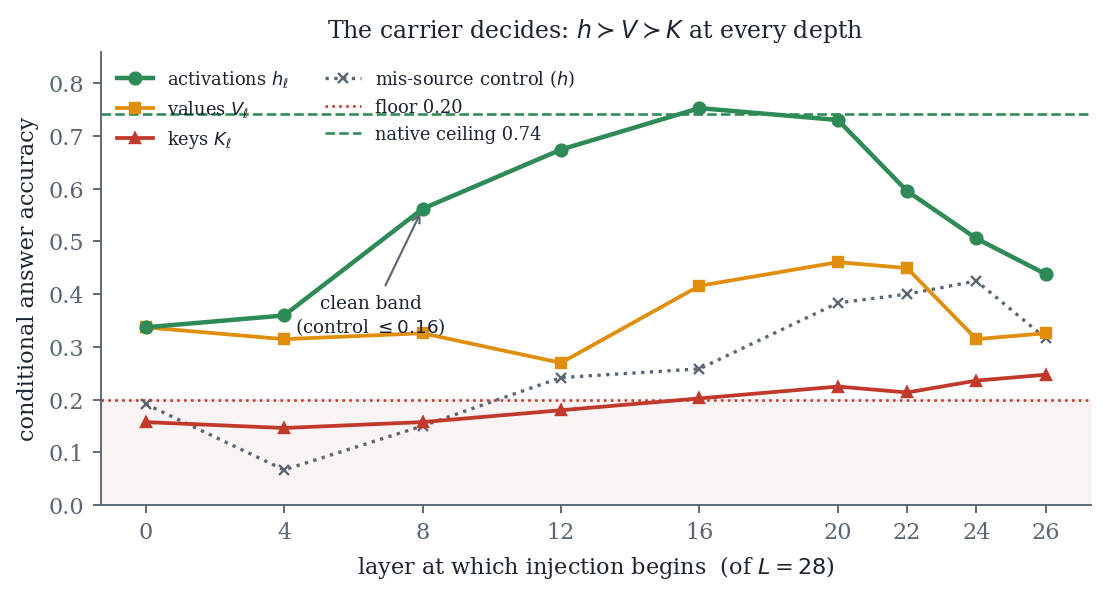}
  \vspace{-10px}
  \caption{Carrier ordering across injection depth: activations dominate values, which dominate keys, and keys never leave the floor band. The clean operating band is shallower than the peak (Section~\ref{sec:acts}).}
  \label{fig:carrier}
\end{figure}

\lead{The carrier ordering.} Holding the protocol fixed and varying only which object is shared gives a depth-stable ordering, activations over values over keys, spanning conditional accuracy $0.753$ down to $0.247$ (Figure~\ref{fig:carrier}). Keys are dead at every depth we probe. The ordering follows the consumption-point argument exactly: the key is read through the query dot-product and softmax, the value is averaged \emph{after} the softmax and so is more forgiving, and the activation is read before either. One qualification. The value carrier's partial transfer appears only in the suffix regime (injecting at every layer from $\ell$ onward); at a single layer, $\Kc$, $\Vc$ and both together are flat-dead (lift $\le +0.025$).

\lead{Why the two carriers differ.} Two alternative explanations would account for the asymmetry without any appeal to the consumption point, and Appendix~\ref{sec:results-mech} rules out both. The wall is not built by amplification: the measured readout gain is $O(1)$ on both sides of the manifold, far too small to turn a tolerable perturbation into a fatal one. It is not built by depth either, since activations injected into the same receiver through the same stack transfer while caches do not. What remains is where each object is read, and the measured tolerances scale with it: the activation carrier has roughly $20\times$ more geometric headroom than the key carrier.

\lead{Where it breaks.} Two boundaries limit every positive result above, and Appendix~\ref{sec:limits} reports both. Cross-family sharing collapses on both carriers: every arm sits at or below its own control, and a frozen projection of foreign keys is dead before any bridge trains, so the bottleneck is representational alignment and not the receiver's adaptability. Separately, the trained arm that first looked strongest on the activation carrier failed a mis-source control, and the arm the controls credit instead scores lower but survives them. A trained sharing mechanism means nothing until it is paired with a mis-source arm and an answer-disjoint split.

\subsection{Summary of the two regimes}

Table~\ref{tab:summary} collects the comparison the paper exists to make. Read by mechanism, the two carriers come out near-mirror images. The cache is dead under substitution and works under augmentation. The activation works under replacement, and works \emph{less} under augmentation.

\begin{table}[t]
\label{tab:summary}
\begin{center}
\begin{tabular}{lll}
\toprule
Mechanism & KV cache ($\Kc,\Vc$) & Activations ($\hs$) \\
\midrule
Substitution / replacement
  & $0.000$ \; \emph{at floor, dead}
  & $\mathbf{0.742}$ \; \emph{at ceiling} \\
Augmentation
  & $\mathbf{0.695}$ \; \emph{at ceiling}
  & $0.438$ \; \emph{above floor, below replacement} \\
\midrule
Winning mechanism & \textbf{augment} & \textbf{replace} \\
Tolerable displacement & $\cos \approx 0.995$ & $\cos \approx 0.90$ \\
Consumption point & through the softmax & before the softmax \\
Cross-family & collapses & collapses \\
\bottomrule
\end{tabular}

\caption{The asymmetry. Best result per (carrier, mechanism) cell, each against its own probe's anchors. Cache numbers are shared-fact accuracy at $n{=}200$ (floor $0.18$, ceiling $0.665$); activation numbers are conditional accuracy at $n{=}120$ (floor $0.20$, ceiling $0.742$). Cells are not comparable across columns in absolute terms, only relative to their anchors.}
\end{center}
\end{table}

\section{Conclusion and Further Work}
\label{sec:conclusion}

This paper asked whether internal state, requiring no model output, can serve as the channel between frozen agents in place of text. The answer depends less on \emph{what} is shared than on how it is consumed. One carrier behaves oppositely under the two mechanisms: the KV cache cannot be substituted, because fidelity and consumability come apart under the softmax that reads it, and the same cache can be augmented. The activation inverts the rule. What unifies them is where the representation enters the receiver's computation. \emph{Augment the cache; replace the activation.}

\lead{Limitations.} Our conclusions are restricted to a controlled latent fact-transfer setting. Only the substitution bound is architecture-general (Appendix~\ref{sec:arch}); every \emph{positive} result uses one model family, so we do not establish that the positive half of our rule holds across architectures, scales or attention shapes, and the mechanism we propose is architectural. The analytic constant $\sqrt{2G}$ is validated at a single group size, which the seven-receiver sweep varies only incidentally alongside depth, width and tokeniser. Cross-family transfer collapses on both carriers under every alignment mechanism we test, and the held-out-sender sweep (Appendix~\ref{sec:heldout}) confines the usable range to a narrow band around the receiver, so our rule holds in a compatibility regime only.

The holder's cache contains the answer verbatim, a standing confound on every cache-side number. The hardened run of Section~\ref{sec:kv} clears the bar, but it shows only that copying does not explain the result, and it covers one arm.

Activation replacement is given oracle position and oracle layer, neither of which a deployed agent knows, so that result speaks only to replacement \emph{correctly aligned}. The pool of Appendix~\ref{sec:multi} is controlled multi-source retrieval and not a multi-agent workflow. Every run uses a single seed, so we claim that query-side augmentation \emph{reaches} the native reference and never that it exceeds it. The accounting in Section~\ref{sec:intro} concerns the number of fitted alignment mechanisms, and says nothing about receiver-side computation, bandwidth, or the memory held by the pool.

\lead{Further work.} \emph{Cross-family alignment} is the blocking problem, and the activation carrier's wider geometric headroom makes it the more promising substrate. The informative experiment compares against closed-form alignment of last-layer states~\citep{statebridge}, which succeeds cross-model where our intermediate-layer injection fails; separating the alignment method from the injection depth would tell us whether the wall is real. Our account also predicts that fragility tracks the sharpness of the consuming readout and not depth, which existing evidence confounds, so varying attention sharpness and depth \emph{independently}, through GQA group size and attention temperature, would test the consumption-point mechanism. Multiple seeds would settle whether query-side augmentation exceeds the native reference.

Nothing blocks internal communication between frozen agents in principle. What blocks it is the framing of \emph{substituting} one agent's state for another's, which recent cache-level messaging proposals adopt~\citep{c2c,latentkv}. Agents do not need to replace one another's internal state; they need to attend over it, or, where the carrier is read before the softmax, to receive it directly.

\subsection*{Use of AI tools}

We disclose the use of generative AI in this work.

A LLM assistant edited and implemented some experiment code, including the manual attention forward pass that permits injection, the metric and control harnesses, and the plotting scripts; it assisted in analyzing and interpreting results; it gave feedback on experimental design, and proposed a few controls, including the mis-source arm and the answer-disjoint split whose application demoted one of our own earlier numbers; and it edited sections of this manuscript and compressed them to the page limit. It also searched for and summarized related work, and helped structure the paper.

We did not use generative AI to generate synthetic data, and the theoretical framing of the readout map and the tolerable set is our own, though an assistant helped state it formally and sharpened two claims we had originally written too strongly.

Every number in this paper comes from a logged run, and the numbers reported here were checked against the raw result files rather than against any assistant's summary of them. Verification gates on the experiment code (parity against native decoding, cache rebuild to $10^{-4}$, and a bit-identical zero-initialized bridge) are themselves automated checks on assistant-written code, and we read their output before reading any result. We reviewed all AI-assisted work and we take responsibility for the final content of this paper, including its text, its claims, and the artifacts produced with AI assistance.

\bibliography{references}
\bibliographystyle{iclr2027_conference}

\appendix
\section{Appendix}

\lead{Conditioning measurement.} The readout Jacobian gain $\kappa$ is estimated over $40$ probe contexts, perturbation magnitudes $\delta \in \{0.005, 0.01, 0.02, 0.05, 0.1, 0.2\}$, $8$ random directions per probe and $3$ seeds, with an on-/off-manifold tolerance of $0.1$. Mean gains are $\kappa_{\mathrm{off}} = 0.506$ and $\kappa_{\mathrm{on}} = 0.343$, a ratio of $1.475\times$. The analytic Jacobian bound predicts the per-layer measured gain at Pearson correlation $0.999$, with governing constant $\sqrt{2G} = \sqrt{12} \approx 3.46$ for $G{=}6$. Per-layer gains are heterogeneous, small early (around $0.11$ at layer 1) and rising to roughly $0.72$ by layer 6, but nowhere exhibit a cliff.

\lead{Bridge training.} Query-side bridges are hand-rolled rank-16 LoRA on the receiver's own $q$, $k$ and $v$ projections at all $28$ layers, with $\Delta = (hA)B$, $A \sim 0.01\cdot\mathcal{N}(0,1)$ and $B$ initialized to zero, so the untrained module is exactly the identity. Training minimizes cross-entropy of the gold answer's first token at a single reserved decision step, through the frozen receiver, with Adam at learning rate $2\times10^{-4}$, gradient clipping $1.0$, and four epochs over $200$ training items; a keep-best guard evaluates a $32$-item slice each epoch. We note as a limitation that this slice is small enough that its scores do not track the $200$-item evaluation, so checkpoint selection is noisy. A clean version of the adaptation ladder needs a proper validation split.

\lead{Substitution sweep.} For completeness, the substitution mechanisms surveyed in Section~\ref{sec:kv} and their transfer against the floor of $0.18$: freeform regression $0.00$; manifold-matched training objective $0.01$; by-construction column-space synthesis $0.01$; row-space target $0.085$; worst-case hinge $0.08$; high-capacity per-layer MLP $0.20$; codebook selection $0.00$; oracle nearest-genuine-key $0.000$. The identity arm reproduces the ceiling $0.665$ exactly. Fidelity within this family is uncorrelated with transfer: the best-fidelity member (codebook selection, cosine $0.880$, attention-logit correlation $0.928$) transfers at zero.

\subsection{Multi-source sharing: a pool of holders}
\label{sec:multi}

A pairwise result is not yet a multi-source one. We study the primitive a multi-agent internal channel would need: sender-agnostic retrieval of task-relevant state from a shared pool. Concretely, we place the receiver in front of a pool of $N$ holders, of which exactly one has read the answer-bearing passage and the other $N{-}1$ have read distractor passages, and let it cross-attend over the union of their caches. The bridge is trained once at pool size $4$ and evaluated at every $N$, so $N \in \{1,2,8\}$ also tests generalization across pool size.

\begin{figure}[t]
  \centering
  \includegraphics[width=0.98\linewidth]{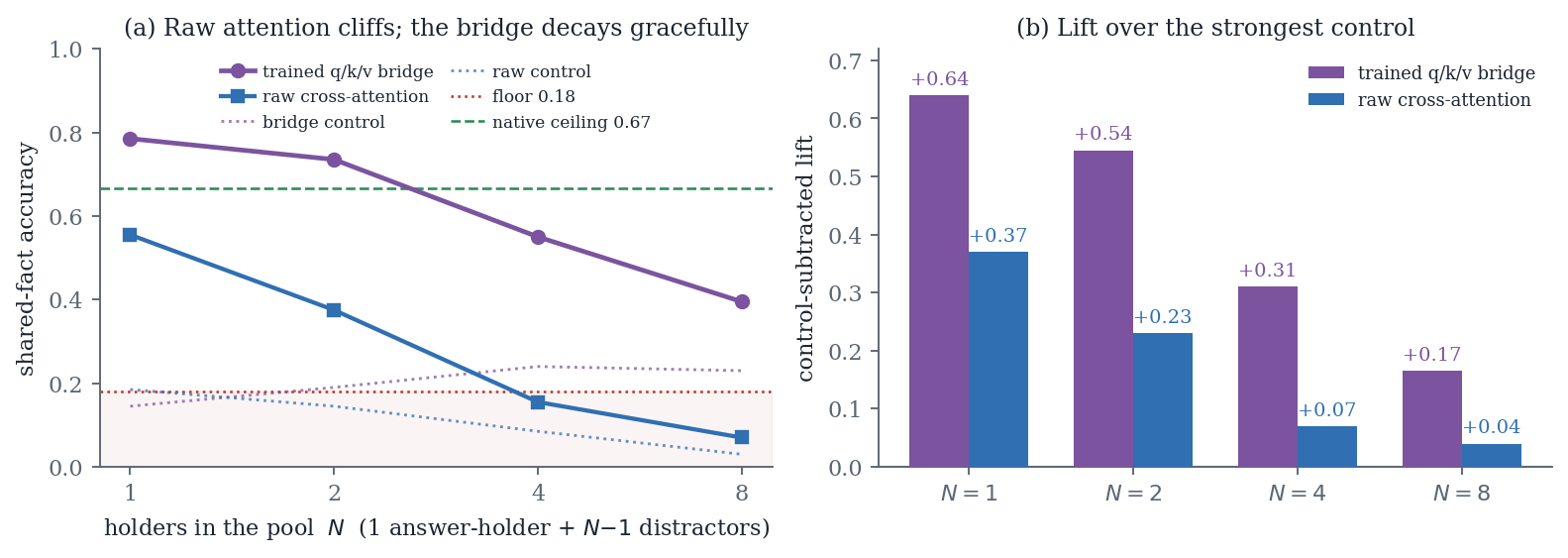}
  \caption{Multi-source KV augmentation over a pool of $N$ holders, one of which read the answer-bearing passage. \textbf{(a)} Raw cross-attention collapses below the floor by $N{=}4$; the trained query-side bridge decays gracefully and remains above floor at $N{=}8$. Dotted lines are the corresponding controls, which themselves lift above floor at large $N$. \textbf{(b)} Lift over the strongest control at each pool size: positive throughout, eroding with $N$.}
  \label{fig:multi}
\end{figure}

Figure~\ref{fig:multi} shows a clean dissociation. Raw cross-attention \emph{cliffs}: $0.555 \to 0.375 \to 0.155 \to 0.070$, falling below the floor of $0.18$ by $N{=}4$. Left to itself, the receiver cannot locate the answer-holder among distractors. The trained query-side bridge \emph{decays gracefully}: $0.785 \to 0.735 \to 0.550 \to 0.395$, still more than twice the floor at $N{=}8$. The query side is doing selection work that raw attention cannot do.

Honesty requires the control-subtracted view, and we lead with it. Both controls (pool with the answer-holder also masked; pool of pure distractors) sit cleanly below floor at $N \le 2$, but climb to $\approx 0.23$--$0.24$ at $N \ge 4$. With a large pool and no answer present, the trained receiver still answers about a quarter of the time by leaning on its own masked prefix and priors. Subtracting the strongest control gives lifts of $+0.640$, $+0.545$, $+0.310$ and $+0.165$ at $N{=}1,2,4,8$. The lift is positive at every pool size, and eroding. That is the substantive finding, and it is smaller than the raw numbers suggest.

\subsection{Held-out senders: how many agents one bridge serves}
\label{sec:heldout}

The pool of Section~\ref{sec:multi} varies what each holder read, not who read it. Every holder there is the receiver's own weights over a different passage, so that experiment establishes invariance to pool size and to the answer-holder's slot. What the sharing argument needs is stronger: a mechanism that does not care which agent published the state. We test that here.

\lead{Senders that differ only in weights.} Holders are drawn from four Qwen2.5-1.5B checkpoints: the receiver itself, the instruction-tuned model, the code model and the math model. All four share the receiver's depth, width and head layout, and their tokenisers are bit-identical, so the comparison isolates sender identity. This matters because a cross-family holder needs a projection between incompatible key spaces, and Section~\ref{sec:limits} shows that the projection is what fails. Here there is no projection to blame. The math checkpoint ships $\theta_{\mathrm{RoPE}}{=}10^4$ against the receiver's $10^6$; keys are stored after rotation, so we force the receiver's $\theta$ on every sender and a frequency mismatch cannot masquerade as a sender effect. Each sender must clear three gates before its numbers are read: the receiver's exact cache shape, finite values, and measurable distinctness from the receiver. The three held-out senders sit at cosine $0.9994$, $0.9933$ and $0.9810$ to the receiver's own cache, which gives a graded axis rather than a binary contrast.

One bridge is trained, on pools built from the receiver's own weights alone, and then evaluated against each sender unchanged. Two cells pin the baseline: raw attention scores $0.155$ and the trained bridge $0.550$, with controls at $0.085/0.045$ and $0.200/0.240$. Both reproduce the $N{=}4$ column of Section~\ref{sec:multi} on all four values, so sender identity is the only quantity that varies between cells.

\lead{The nearest sender transfers; the next does not.} Table~\ref{tab:heldout} gives the result. At cosine $0.9994$ the bridge reaches $0.550$ against a strongest control of $0.225$, a control-subtracted lift of $+0.325$ with a $95\%$ interval of $[+0.235,+0.415]$. That equals what the bridge achieves on the sender it trained on, and it is earned rather than inherited: raw attention on the same sender reaches only $0.145$, below the floor of $0.18$. The adapter is doing sender-agnostic work. One step further out the effect is gone. At cosine $0.9933$ the lift is $+0.030$ with an interval of $[-0.032,+0.092]$, and at $0.9810$ it is $-0.005$. Neither is distinguishable from its own control, and both sit below the floor in absolute terms.

\begin{table}[t]
\label{tab:heldout}
\centering
\small
\begin{tabular}{llccccc}
\toprule
sender & cosine & rel.\ diff. & raw lift & bridge shared & bridge lift & $95\%$ interval \\
\midrule
self (seen)        & ---      & ---      & $+0.070$ & $0.550$ & $+0.310$ & $[+0.219,+0.401]$ \\
instruct (held out)& $0.9994$ & $0.033$  & $+0.070$ & $0.550$ & $+0.325$ & $[+0.235,+0.415]$ \\
coder (held out)   & $0.9933$ & $0.170$  & $+0.000$ & $0.130$ & $+0.030$ & $[-0.032,+0.092]$ \\
math (held out)    & $0.9810$ & $0.433$  & $+0.000$ & $0.100$ & $-0.005$ & $[-0.064,+0.054]$ \\
\bottomrule
\end{tabular}
\caption{Held-out senders. One rank-16 query-side bridge, trained on pools built from the receiver's own weights only, evaluated against senders it never saw. Senders are Qwen2.5-1.5B checkpoints with the receiver's geometry and a bit-identical tokeniser, ordered by the cosine between their cache and the receiver's. We report relative Frobenius difference alongside cosine: the two rank the senders identically, so the ordering is not an artifact of one metric, though two metrics over three checkpoints cannot settle which is the right notion of compatibility. Lift is shared accuracy minus the stronger of the two controls; intervals are $95\%$ normal-approximation intervals on that difference. Pool size $N{=}4$, $n{=}200$, floor $0.18$, native ceiling $0.665$. The \emph{self} row reproduces the $N{=}4$ point of Figure~\ref{fig:multi} exactly.}
\end{table}

\lead{Where the radius comes from.} The receiver's tolerance for a displaced key is narrow and is not our finding: it is the tolerable set of Section~\ref{sec:theory}, and the displacement at which keys stop being readable has been characterized independently. What this experiment adds is that the same narrowness governs \emph{who} may join a pool, not only \emph{what} may be substituted into one. Augmentation lets the receiver keep its own cache, so it escapes the bound for a cache the receiver could have produced; it does not repeal the bound for a cache produced by different weights. The radius is therefore a property we inherit, and the contribution here is measuring where it falls on the sender axis.

\lead{What this buys.} The sharing argument of Section~\ref{sec:intro} needs one mechanism that serves every publisher, and a pairwise bridge per ordered pair is what it replaces. The measurement says how far that replacement goes. Within a narrow band of the receiver, one bridge covers senders it never trained on, so such a pool needs one fitted bridge rather than one per pair. Outside the band it covers nothing, and those agents fall back to text or to a fitted pair. Sharing is therefore a real alternative to transfer for a population of closely related agents, which is the common case when a deployment scales one base model into differently tuned roles, and it is not yet an alternative for a heterogeneous population. We state the bounded version because it tells a system designer which agents may share a pool.

\begin{figure}[t]
  \centering
  \includegraphics[width=0.98\linewidth]{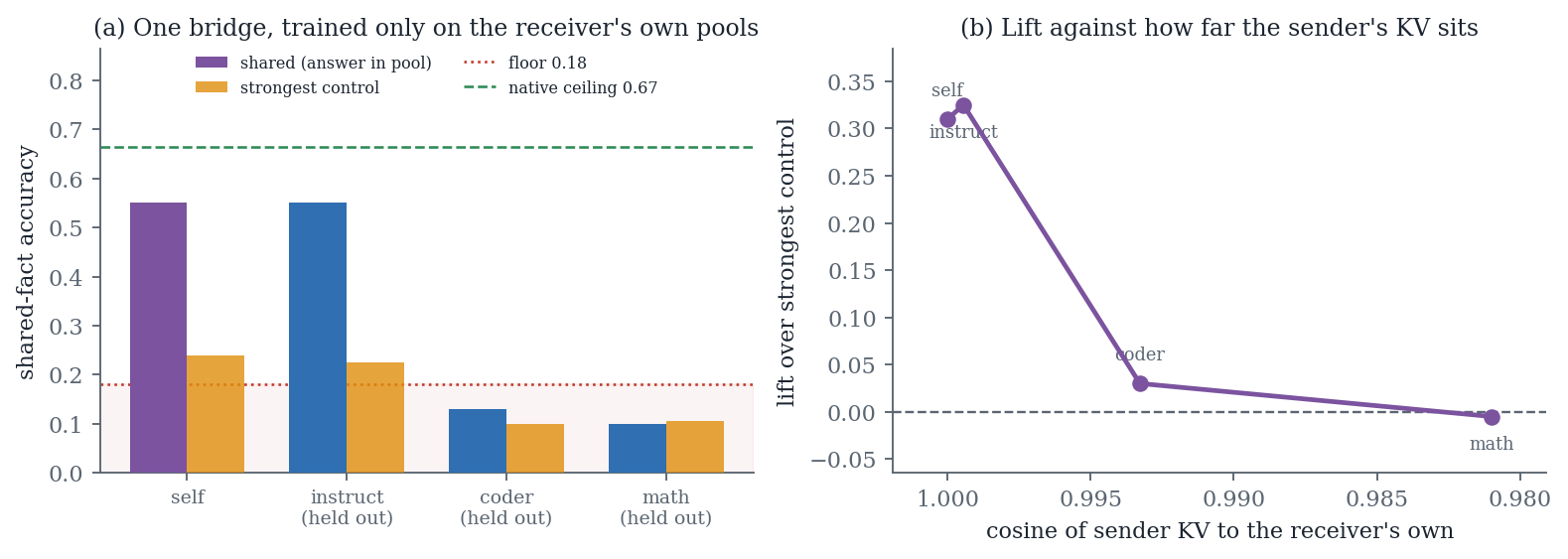}
  \caption{Held-out senders. \textbf{(a)} The bridge, trained on the receiver's own pools only, reaches the same accuracy on the nearest held-out sender as on the sender it trained on, and falls to its control on the two more distant senders. \textbf{(b)} Control-subtracted lift against the cosine between the sender's cache and the receiver's own. The effect is present at $0.9994$ and absent by $0.9933$, which measures where the effect falls on the sender axis. Three checkpoints locate a transition, not a radius.}
  \label{fig:heldout}
\end{figure}

\subsection{Does the wall depend on the architecture?}
\label{sec:arch}

Everything above holds one receiver fixed, which leaves open whether the substitution bound is a property of attention or a property of Qwen2.5-1.5B. The mechanism we propose is architectural, so this is a question the paper has to answer rather than defer. We repeat the oracle-substitution probe of Section~\ref{sec:kv} on seven receivers spanning three families, six depths and five attention geometries, re-deriving each receiver's own floor and ceiling so that every row is read against its own anchors.

\begin{table}[t]
\label{tab:arch}
\centering
\small
\begin{tabular}{lccccc}
\toprule
receiver & $L$ & $G$ & floor & identity & nearest genuine \\
\midrule
Pythia-70M & $6$ & $1$ & $0.000$ & $0.025$ & $0.000$ \\
GPT-2 & $12$ & $1$ & $0.033$ & $0.233$ & $0.000$ \\
Pythia-410M & $24$ & $1$ & $0.067$ & $0.375$ & $0.000$ \\
Pythia-1.4B & $24$ & $1$ & $0.092$ & $0.517$ & $0.000$ \\
Qwen2.5-0.5B & $24$ & $7$ & $0.092$ & $0.633$ & $0.008$ \\
Qwen2.5-1.5B & $28$ & $6$ & $0.200$ & $0.742$ & $0.000$ \\
Qwen2.5-3B & $36$ & $8$ & $0.350$ & $0.792$ & $0.000$ \\
\bottomrule
\end{tabular}
\caption{The substitution bound across architectures. Each receiver is probed with its own bank of genuine keys and its own anchors. \emph{Identity} hands the receiver its own exact keys and recovers that receiver's ceiling in every case, which verifies the harness on each architecture. \emph{Nearest genuine} substitutes the closest key the receiver itself produced elsewhere. Depth spans $L{=}6$ to $L{=}36$ and the grouped-query factor spans $G{=}1$ (multi-head) to $G{=}8$. $n{=}120$.}
\end{table}

Table~\ref{tab:arch} gives the result. On every receiver the identity arm reproduces that receiver's ceiling exactly, so the manual injection path is faithful on all seven architectures. And on every receiver the nearest genuine key transfers at or below $0.008$, against floors ranging from $0.000$ to $0.350$. The bound is not a Qwen property. It appears at six layers and at thirty-six, in three model families, and under both multi-head attention ($G{=}1$) and grouped-query attention up to $G{=}8$.

Two qualifications matter. First, the variation in $G$ here is incidental rather than controlled: these checkpoints differ in depth, width, training data and tokeniser at the same time, so the table shows that the bound survives architectural variation without isolating $G$ as the variable responsible. The clean experiment for the $\sqrt{2G}$ prediction of Section~\ref{sec:theory} would vary group size within one architecture, and we have not run it. Second, this sweep probes substitution only. It establishes that the failure mode generalizes; it does not establish that the positive half of our rule, augmentation on the cache and replacement on the activation, generalizes with it. The receivers here are also probed in isolation rather than paired, so nothing in this table speaks to cross-family sharing, which Section~\ref{sec:limits} reports separately as a collapse.

\subsection{Methodological detail: model geometry and verification gates}
\label{app:methods-detail}

\lead{Receiver and holder geometry.} The frozen Qwen2.5-1.5B~\citep{qwen2.5} receiver
has $L{=}28$ layers and hidden width $1536$, with grouped-query attention over
$N_Q{=}12$ query heads and $N_{KV}{=}2$ key/value heads ($G{=}6$), head dimension
$128$, and key/value width $256$. The cross-family holder, GPT-Neo-1.3B, has $24$
layers and key/value width $2048$. We project it into the receiver's key space with
a frozen per-layer SVD map fitted on a $40$-item sample, aligning depth
proportionally between the two stacks.

\lead{Verification.} All attention is re-implemented manually to permit injection, so
every run passes two gates before its numbers are read. A \emph{parity} gate requires
the manual forward to reproduce native greedy decoding string-exactly from the same
prefix. A \emph{rebuild} gate requires reconstructed prefix keys and values to match
the native cache (maximum absolute deviation $\le 10^{-4}$). Trained arms additionally
assert that the zero-initialized bridge scores bit-identically to the untrained arm at
epoch zero.

\subsection{Mechanism: the wall is at the readout, not the depth}
\label{sec:results-mech}

\begin{figure}[t]
  \centering
  \includegraphics[width=0.98\linewidth]{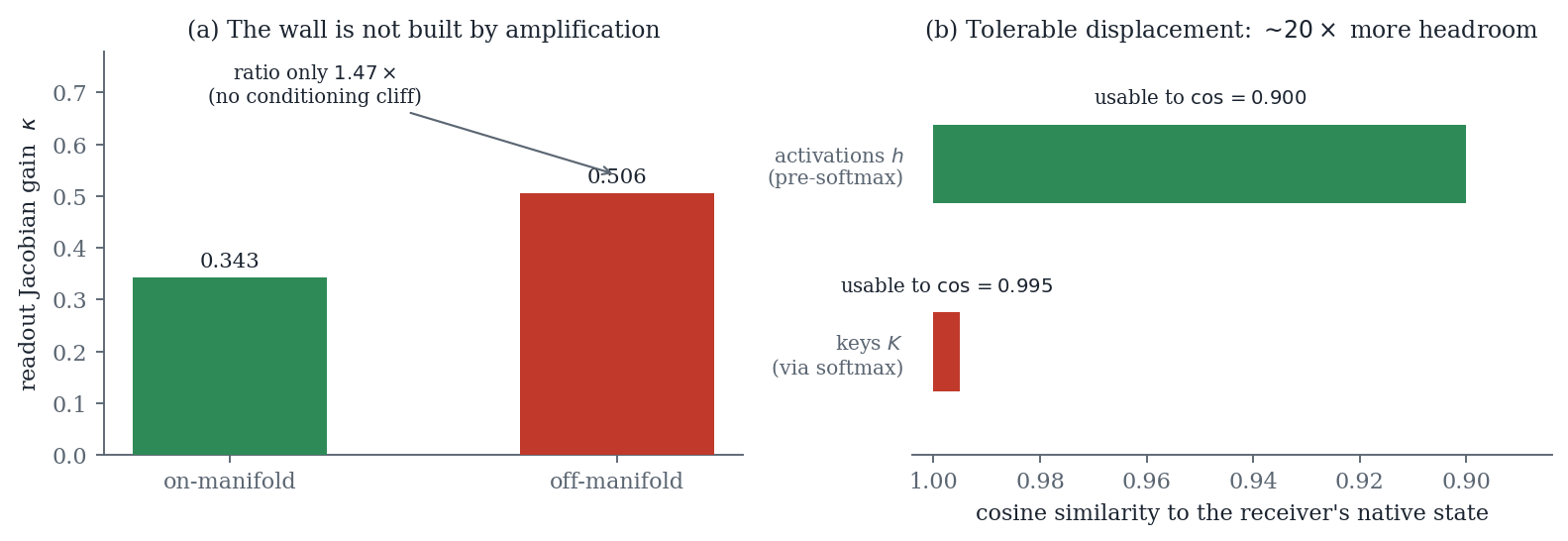}
  \caption{Why the two carriers differ. \textbf{(a)} The measured readout Jacobian gain is $O(1)$ both on and off manifold, with a ratio of only $1.47\times$: the wall is not built by amplification, so no single layer and no stacking of layers blows up a tolerable perturbation. \textbf{(b)} Tolerable displacement by carrier: activations remain usable to cosine $0.90$, keys only to $\approx 0.995$, about $20\times$ less headroom. Same receiver, same depth; the difference is the consumption point.}
  \label{fig:mech}
\end{figure}

Proposition~\ref{prop:notdepth} makes two claims we can measure. The first is that there is no conditioning cliff. Estimating the readout Jacobian gain over $40$ probes, six perturbation magnitudes, eight directions per probe and three seeds gives $\kappa_{\mathrm{off}} = 0.506$ against $\kappa_{\mathrm{on}} = 0.343$, a ratio of $1.47\times$ (Figure~\ref{fig:mech}a). Off-manifold displacement is modestly more damaging, not catastrophically so. Per-layer gains are $O(1)$ on both sides and no single layer explodes. An analytic Jacobian bound predicts the measured per-slot gain at Pearson correlation $0.999$ with the governing constant $\sqrt{2G}$. This pins the sensitivity to the shape of attention, and it is far too small to be the wall.

The second claim is that depth is not the culprit. The control is the carrier comparison itself: activations injected into the identical $28$-layer receiver, consumed through the identical stack, transfer at $0.72$ while caches do not. If per-layer error compounded to unreadability, the activation would die alongside the cache. It does not. What differs is only \emph{where} in each layer the object is consumed.

The tolerance measurements close the argument quantitatively (Figure~\ref{fig:mech}b). Perturbing a shared activation and sweeping matched cosine, conditional accuracy stays near $0.70$ all the way down to cosine $0.90$. The KV carrier is at the floor across the same sweep, and genuine nearest-neighbor keys at cosine $\approx 0.995$ already fail. That is roughly $20\times$ more geometric headroom for the activation, the budget a learned cross-model bridge would have to spend.

\subsection{Where it breaks: cross-family transfer, and a control that bites}
\label{sec:limits}

Two boundaries limit every positive result above, and both deserve to be stated plainly.

\lead{Cross-family collapses on both carriers.} With a GPT-Neo-1.3B holder and a Qwen receiver, every KV augmentation arm sits at or below its own control (Table~\ref{tab:xkv}, lower block; the best is $0.080$ against a control of $0.085$, i.e. noise). The identity sanity arm is $0.000$: the frozen SVD projection of foreign keys is dead before any bridge trains. On the activation side, all eight cross-family cells of the trained-router experiment also sit at floor, with the trained arm scoring \emph{identically} to the uniform one, so the bridge learned nothing usable. Rank-16 adaptation of the receiver's own projections does not bridge a foreign representation. The bottleneck here is representational alignment, rather than the receiver's adaptability or the carrier.

\begin{figure}[t]
  \centering
  \includegraphics[width=0.98\linewidth]{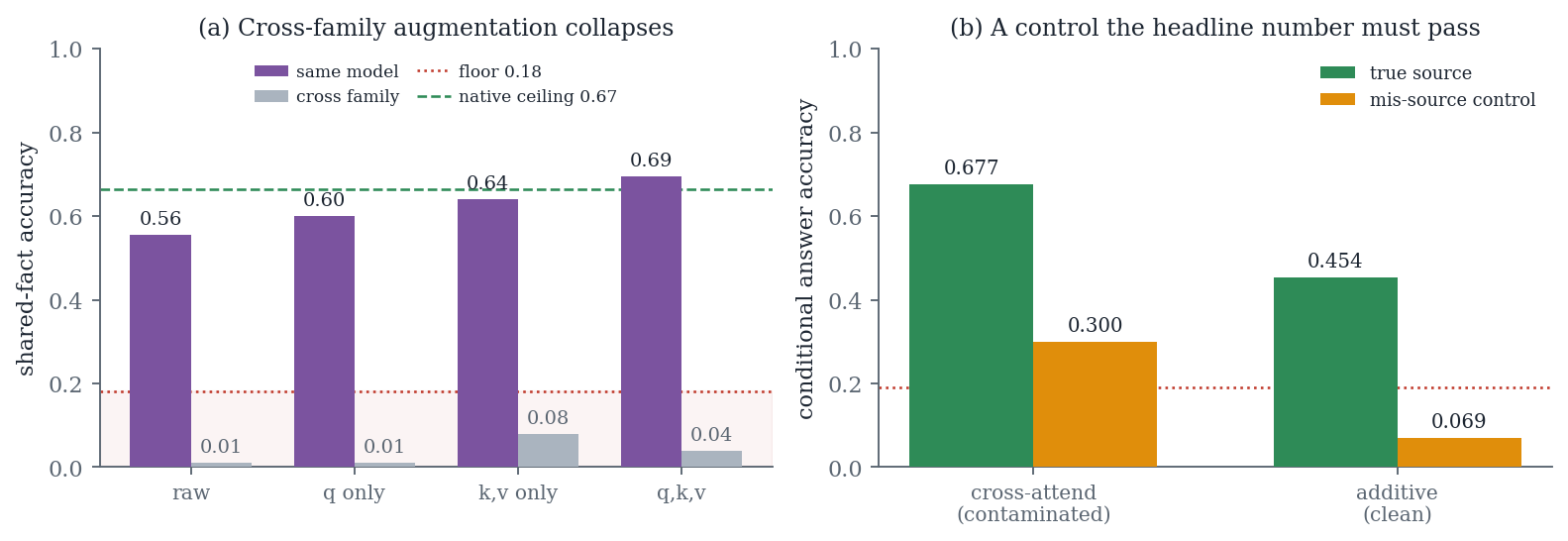}
  \caption{The boundaries. \textbf{(a)} Cross-family KV augmentation collapses at every rung of the adaptation ladder, while the same ladder within one model reaches the native ceiling. \textbf{(b)} On the activation carrier, the apparently strongest trained arm fails a mis-source control ($0.300$ against a floor of $0.19$); the additive arm passes it ($0.069$) at a lower but genuine $0.454$.}
  \label{fig:limits}
\end{figure}

\lead{A trained-router positive, and the control that halved it.} Training a router on the activation carrier initially produced a striking number: conditional accuracy $0.677$ for a trained cross-attention consumption path, against a uniform-router baseline at floor. A mis-source control and an answer-disjoint split, run afterwards, showed this to be substantially contaminated. Feeding the router activations from a \emph{different} item still yielded $0.300$, far above the floor of $0.19$, so a large part of the headline was the router emitting answers regardless of source. The same controls credit a different arm (additive consumption, router-only adaptation) with conditional accuracy $0.454$ against a mis-source control of $0.069$, a genuine controlled positive on unseen answers. Both LoRA-adapted arms, meanwhile, collapse to $\le 0.115$ once answers are disjoint, which reveals them as answer-vocabulary memorisation (Figure~\ref{fig:limits}). We report the $0.454$ as the defensible trained-router result and treat the $0.677$ as an artifact. The transferable lesson is that a trained sharing mechanism must be paired with a mis-source arm and an answer-disjoint split before its number means anything.

\lead{Controls in full.} \emph{No-op}: inject the receiver's \emph{own} masked state at the same positions and layers, separating ``something was overwritten here'' from ``the holder's content was overwritten in.'' \emph{Position}: place the holder's state at the fact-free common prefix. \emph{Mismatch / mis-source}: supply the holder's state for a \emph{different} item, which must collapse if the mechanism consumes source content instead of emitting memorised answers. \emph{Answer-disjoint split}: for trained bridges, no normalized eval answer appears anywhere in train.

\end{document}